\documentclass{article}

\usepackage{latexsym}
\usepackage{amssymb}
\usepackage{amsmath}
\usepackage{amsthm}
\usepackage{algorithm}
\usepackage{algorithmic}
\usepackage{booktabs}
\usepackage{enumitem}
\usepackage{graphicx}
\usepackage{color}
\usepackage{xcolor}
\usepackage{kotex}
\usepackage{multirow}
\usepackage{PRIMEarxiv}
\usepackage[utf8]{inputenc} 
\usepackage[T1]{fontenc}    
\usepackage{hyperref}       
\usepackage{url}            
\usepackage{booktabs}       
\usepackage{amsfonts}       
\usepackage{nicefrac}       
\usepackage{microtype}      
\usepackage{lipsum}
\usepackage{fancyhdr}       
\usepackage{graphicx}       
\graphicspath{{media/}}     

\title{Personalized Federated Learning via Sequential \\Layer Expansion in Representation Learning
}

\author{
  Jaewon Jang, Bonjun Choi \\
  Computer Science and Engineering \\
  Soongsil University \\
  Seoul, South Korea\\
  \texttt{\{jwon0524, davidchoi\}@soongsil.ac.kr} \\
}

\begin{document}
\maketitle

\begin{abstract}
Federated learning ensures the privacy of clients by conducting distributed training on individual client devices and sharing only the model weights with a central server. However, in real-world scenarios, the heterogeneity of data among clients necessitates appropriate personalization methods. In this paper, we aim to address this heterogeneity using a form of parameter decoupling known as representation learning. Representation learning divides deep learning models into 'base' and 'head' components. The base component, capturing common features across all clients, is shared with the server, while the head component, capturing unique features specific to individual clients, remains local. We propose a new representation learning-based approach that suggests decoupling the entire deep learning model into more densely divided parts with the application of suitable scheduling methods, which can benefit not only data heterogeneity but also class heterogeneity. In this paper, we compare and analyze two layer scheduling approaches, namely forward (\textit{Vanilla}) and backward (\textit{Anti}), in the context of data and class heterogeneity among clients. Our experimental results show that the proposed algorithm, when compared to existing personalized federated learning algorithms, achieves increased accuracy, especially under challenging conditions, while reducing computation costs.
\end{abstract}



\section{Introduction}

Traditional centralized machine learning collects all raw data to a central server for training. However, concerns about data privacy have grown among companies and researchers in the era of big data. Moreover, the implementation of stringent legal regulations, such as the General Data Protection Regulation (GDPR)~\cite{jacobs1963fine} in Europe and the California Consumer Privacy Act (CCPA)~\cite{mactaggert2019california} in the United States, has underscored the need for new methods of collecting and sharing client data that comply with these laws. In this context, federated learning has emerged as a significant research area, allowing the utilization of diverse data distributed across various devices like smartphones, IoT devices, and wearables while ensuring data privacy.


Federated learning~\cite{mcmahan2017communication}, in contrast to centralized machine learning, does not directly share raw client data with a central server. Instead, it trains models on each edge device and shares only the model's weights, aggregating them to update a global model. This training approach ensures privacy, as raw data is not shared with the server, and it offers communication cost advantages when dealing with large datasets. However, in real-world scenarios, local data among clients can be highly heterogeneous, exhibiting varying data distributions. This heterogeneity can lower the performance of the global model during server-side aggregation and may introduce bias towards certain clients.



Therefore, the need for suitable personalization methodologies arises to capture and optimize each client's unique data characteristics. To address these issues, personalized federated learning (PFL) approaches~\cite{li2020federated,karimireddy2020scaffold} are actively being researched. In this study, we utilize representation learning which is one of the PFL approaches. Representation learning divides the layers of a deep learning model into \textit{base} and \textit{head} components. In representation learning, the \textit{base (or feature extractor)} part represents common features for all clients and is shared with the server and all clients. The \textit{head (or classifier)} part represents unique features for specific clients and remains on the local device, not shared with the server. Typically, the \textit{head} part uses the last fully connected layer (classification layer), and in our experimental setup, we also designate the \textit{head} part as the last fully connected layer while setting the remaining layers as the \textit{base} part.


Our idea originated with the goal of subdividing the components of the deep learning model in representation learning more densely than just the base and head. Additionally, the layer scheduling approach drew inspired from curriculum learning~\cite{bengio2009curriculum}, resulting in the development of both Vanilla and Anti scheduling.


Curriculum learning is a method that, similar to how human learning, gradually progresses from easy examples to more challenging ones. 
Traditional curriculum learning focuses on determining the difficulty of datasets and utilizes pre-trained expert models on the entire dataset to evaluate the difficulty of each data point. Subsequently, it assigns difficulty scores based on the client's loss using the pre-trained expert model and then conducts training either by starting with easy examples and progressing to harder ones or vice versa. In the recently researched federated learning environment, curriculum learning~\cite{vahidian2023curricula} also necessitates a well-trained expert model.


However, curriculum learning relies on expert models to assess the difficulty of data points. This dependence can result in significantly reduced performance if the expert models are compromised by malicious attacks or poorly trained. Furthermore, the process of assigning difficulty scores and dividing datasets based on difficulty can complicate operations in a federated learning environment, as highlighted by ~\cite{vahidian2023curricula}. In contrast to curriculum learning, our scheduling algorithm does not categorize data based on difficulty levels. Instead, it focuses on decoupling layers according to the principles of representation learning. In typical deep learning models, the initial layers are responsible for extracting low-level features, while the later layers handle the extraction of more complex and abstract features. Building on this insight, we have concentrated on more densely separating the base layer in representation learning algorithms and sequentially training it, deviating from the traditional curriculum learning approach.


Hence, our algorithm densely divides the base layer, initially freezing the entire layer set, and then progressively unfreezes specific layers for training according to a schedule. There are two scheduling methods: \textit{Vanilla} scheduling, which starts by unfreezing layers closest to the input and progresses towards the output, and \textit{Anti} scheduling, which begins by unfreezing layers nearest to the output and works backward. A key advantage of our proposed algorithm is that, during the early rounds, only a portion of the unfrozen base layer is shared with the server, rather than the entirety. This approach not only enhances performance but also reduces the communication and computational costs compared to other algorithms.


The contributions of this paper are as follows:
\begin{itemize}
    \item Our algorithm densely divides the base layer to address the heterogeneity of the client's data and class distribution, and it proposes two scheduling methods.
    \item The implementation of scheduling reduces the need to communicate all base layers in the early stages of training, thereby cutting down on communication and computational costs. 
    \item In scenarios with both data and class heterogeneity, the \textit{Anti} scheduling approach outperforms other algorithms in terms of accuracy, while the \textit{Vanilla} scheduling method significantly reduces computational costs compared to other algorithms.
    \item We visually present the accuracy for each client and mathematically compare the computational costs of each algorithm.
\end{itemize}

\section{Related Work}
\subsection{Federated Learning}
Federated Learning is a machine learning technique that enables training while ensuring the privacy of clients' distributed data~\cite{mcmahan2017communication}. However, a significant challenge in real-world federated learning scenarios is the heterogeneity of the data distribution among distributed clients, necessitating the application of appropriate personalization techniques.

\subsection{Personalized Federated Learning}
Personalized Federated Learning (PFL) has been proposed to tackle the heterogeneity of client data and can be categorized into several approaches such as Clustering, Meta Learning, and Representation~\cite{tan2022towards}. This section compares three related personalization methods, including representation learning, which are considered similar.

\subsubsection{Meta Learning}
Meta learning, also referred to as learning to learn, was initially proposed for few-shot learning. \cite{finn2017model} focuses on finding model initialization methods that can rapidly adapt to new and unseen tasks, allowing for the quick learning of new tasks with limited data. \cite{nichol2018first} employs first-order algorithms, which are computationally less complex compared to other meta-learning algorithms, and demonstrates comparable performance without the need for second-order methods.


\subsubsection{Transfer Learning}
Transfer learning has also been proposed for few-shot learning and is a method to apply knowledge learned in a specific domain to a different domain. It involves adapting a pre-trained backbone from a different domain to a similar but distinct domain using the well-known method of fine-tuning. \cite{li2019fedmd} helps solve heterogeneity by transferring and learning knowledge between models with different structures. Additionally, \cite{chen2020fedhealth} proposes a federated transfer learning framework for wearable health data.



\subsubsection{Multi-task Learning}
Multi-task learning is a method that can enhance a model's generalization ability by simultaneously learning multiple tasks and sharing knowledge or representations between these tasks. In~\cite{smith2017federated}, a novel approach is proposed that combines federated learning with multi-task learning. This approach improves the performance of each task by addressing multiple related learning tasks simultaneously, enabling efficient learning through the utilization of shared knowledge.


\subsubsection{Representaion Learning}
The PFL technique we used is a kind of representation learning, which is a parameter decoupling method to alleviate statistical heterogeneity. Representation learning is a method of training deep learning models by dividing the layers into a base and a head. The base shares common features among clients, while the head layer keeps in locally and is used for personalization ~\cite{arivazhagan2019federated}.

Moreover, \cite{liang2020think} argues that it generalizes more easily to new devices compared to other learning mechanisms and that fair representations, which obscure protected attributes, are effectively learned through adversarial training. Additionally, \cite{collins2021exploiting} demonstrates fast convergence in linear regression problems by leveraging distributed computational power among clients to perform many local updates. Furthermore, \cite{chen2021bridging} shows that there is a discrepancy in the validation methods between general federated learning algorithms and personalized federated learning algorithms. They argue that both local and global model accuracy should be considered and introduce a new loss function to mitigate class imbalance issues. In addition, \cite{oh2021fedbabu} suggests that updating the head in cases where client data is highly heterogeneous can negatively affect personalization. Therefore, during training rounds, a randomly initialized head is used, employing only the body for training and aggregation. We adopt a similar training setup \cite{oh2021fedbabu}, using a randomly initialized head during training rounds, and the head is only utilized in the final fine-tuning after the training rounds are completed.


\section{Proposed Algorithm}

\begin{table}[htb!]
    \caption{List of Symbols Used}
    \centering
    \begin{tabular}{c|p{7cm}}
    \hline
    Symbol & Description \\
    \hline
    $N$ & Total number of clients \\
    $C_i$ & Client $i$, $i \in \{1,2,\ldots, N\}$ \\
    $|D_i|$ & Number of data points for client $i$ \\
    $|D|$ & Total number of data points \\
    $T$ & Total global rounds\\ 
    $F$ & Fine-tuning rounds \\
    $K$ & Total number of base layers \\
    $t_k$ & Unfreeze point (round) of $k$-th layer  \\
    $\eta$ & Learning Rate \\
    $\mathcal{L}$ & Loss Function \\
    $\theta_i$ & Local parameter for client $i$ \\
    $\theta_G$ & Global model parameter \\
    $\theta_{i,b}$ & Base parameter shared among all clients \\
    $\theta_{i,p}$ & Head parameter of client $i$ \\
    
    \hline
    \end{tabular}
    \label{tab:notations}
\end{table}

\begin{figure}[htb!]
\centering
\includegraphics[width=0.535\textwidth]{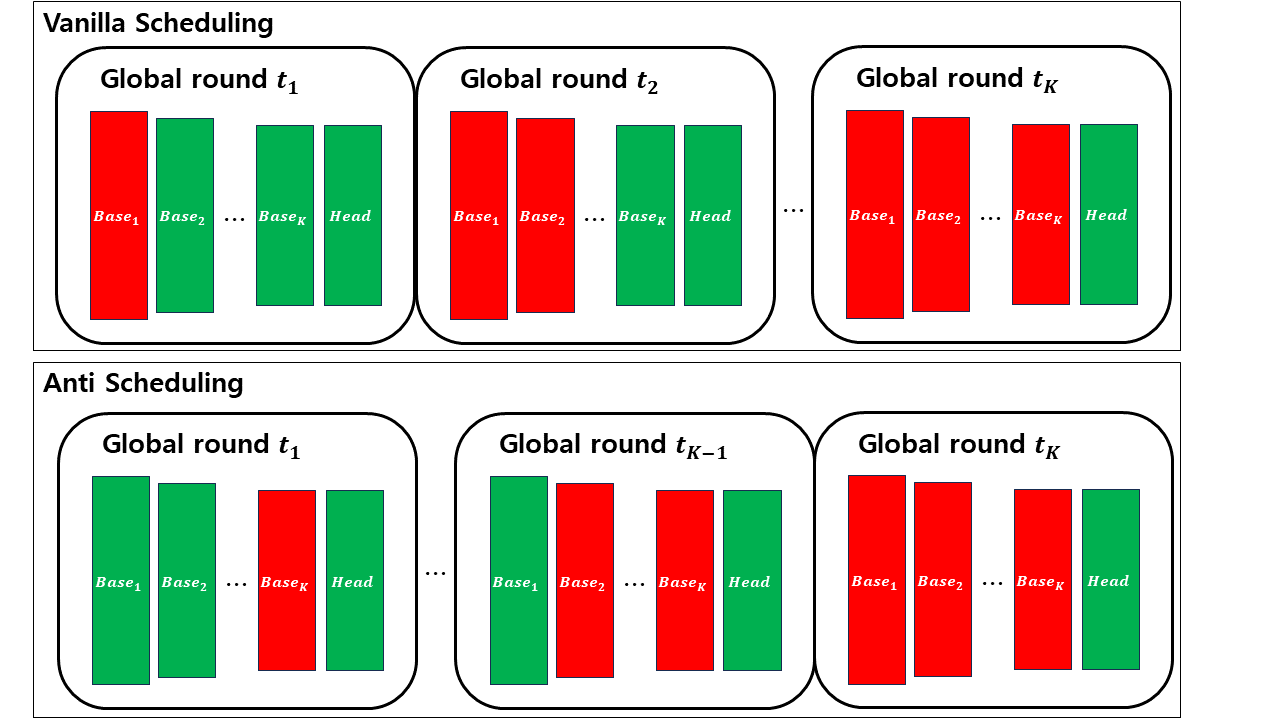}
\caption{Illustration of the proposed Vanilla and Anti Scheduling algorithms. The upper shows Vanilla Scheduling, starting with the shallowest layer and advancing to deeper ones. The lower shows Anti Scheduling, which begins with the deepest layer and progresses inversely.}
\label{vanilla_anti}
\end{figure}

In this section, we present our proposed algorithms in detail, namely \textit{Vanilla Scheduling} and \textit{Anti Scheduling}. As shown in Figure~\ref{vanilla_anti}, Vanilla Scheduling starts by unfreezing the shallowest layers (closest to the input) and progressively moves towards deeper layers. This method allows the model to capture low-level features early in the training process, which is crucial for building a strong foundational understanding before more high-level features are introduced. Conversely, Anti Scheduling begins with the deepest layers (closest to the output) and progressively unfreezes towards the input. This approach prioritizes the learning of high-level features from the outset, which can be beneficial for complex pattern recognition tasks that depend heavily on such features.


As shown in Figure~\ref{vanilla_anti}, the structured progression from shallow to deep layers in Vanilla Scheduling, and the reverse in Anti Scheduling, structurally differ from traditional representation learning.  In traditional representation learning methods, the base and head of the model are typically trained simultaneously without specific layer prioritization. In our approach, however, each layer's training is specifically timed and prioritized. This strategic scheduling enhances the model's ability to adapt to varying complexities of features throughout the training process.




Furthermore, as detailed in Section~\ref{computation cost}, Vanilla Scheduling can achieve comparable accuracy with significantly lower computational costs in environments characterized by high data and class heterogeneity. In contrast, Anti Scheduling excels in achieving the highest accuracy in scenarios where both data and class heterogeneity are pronounced. They both follow the basic federated learning setup and the representation learning as explained below.

As a basic federated learning setup, we assume that each client $C_i$ possess data $\mathbf D_i= \textbf({x}_i,y_i) \in R^d$, where $i \in {1,2,\ldots, N}$ represents $i$-th client out of a total of $N$ clients and $d$ represents the input dimension. Each client $i$ updates its local model parameter $\theta_i$ based on its data $D_i$ and the global model parameter $\theta_G$ as

\begin{equation}
    \theta_i^{(t+1)} = \theta_i^{(t)} - \eta \nabla_{\theta_i} \mathcal{L}(D_i,\theta_i^{(t)}),
\label{eq1}
\end{equation}
where $\eta$ is the learning rate, $\nabla L$  is the gradient of the loss function $L$, and $t \in (1,2,\ldots,T)$ denotes the number of rounds.
The central server updates the global parameter $\theta_G$ based on all clients' local parameter updates as

\begin{equation}
    \theta_G^{(t+1)} = \sum_{i=1}^{N} \frac{|D_i|}{|D|} \theta_i^{(t+1)},
\label{eq2}
\end{equation}
where $|D_i|$ is the number of data points for client $i$, and $|D|$ is the total number of data points.

For representation learning, the model parameters are divided into two components $\theta_i = (\theta_{i,b}, \theta_{i,h})$. Here, $\theta_{i,b}$ represents the base parameter shared among all clients and $\theta_{i,h}$ represents the head parameter of the $i$-th client. The model parameters from Eq.~\eqref{eq1} is modified as follows:

\begin{equation}
    \theta_i^{(t+1)} = \theta_i^{(t)} - \eta \nabla_{\theta_i} \mathcal{L}(D_i, \theta_{i,b}^{(t)}, \theta_{i,h}^{(0)}).
\label{eq3}
\end{equation} 

Excluding FedBABU \cite{oh2021fedbabu}, previous works the gradient of the head parameter is stopped during training and during the aggregation phase. The global model is updated using both the base and head parameters. However, FedBABU and our scheduling algorithm perform both training and aggregation using only the base without involving the head. Only after training, a few rounds of fine-tuning are included using both the base and head.

The aggregation process, which excludes the use of the head, proceeds as follows:


\begin{equation}
    \theta_G^{(t+1)} = \sum_{i=1}^{N} \frac{|D_i|}{|D|} \theta_{i,b}^{(t+1)}.
\label{eq4}
\end{equation}

Our algorithm follows the same training setup as FedBABU~\cite{oh2021fedbabu}, wherein the head layer is not trained or aggregated during training; instead, it is fine-tuned for each client after training completion. In FedBABU, the training round sets the learning rate of the head to zero, allowing gradient computation but preventing its application. In contrast, our scheduling algorithm ensures complete decoupling by freezing gradients in both the head and base parts, preventing both computation and application.

\subsection{Method 1: Vanilla Scheduling}
Vanilla scheduling is a training method that starts by thoroughly learning the shallowest layers (closest to the input) in the base layer, freezing the rest layers, and then progressively unfreezing them. This enables the model to preferentially learn low-level features and patterns, aiding in better understanding the abstract characteristics and patterns of the data. The training steps of Vanilla scheduling, which starts with the shallowest layers, can be represented as


\begin{small}
\begin{align}
            \theta_i^{(t_1+1)} = \theta_i^{(t_1)} - \eta \nabla_{\theta_i} \mathcal{L}(D_i, &\theta_{i,b_{1}}^{(t_1)}, \theta_{i,b_{2}}^{(0)}, \ldots , \theta_{i,b_{K-1}}^{(0)}, \theta_{i,b_{K}}^{(0)}, \theta_{i,h}^{(0)})  
            \nonumber \\
            \theta_i^{(t_2+1)} = \theta_i^{(t_2)} - \eta \nabla_{\theta_i} \mathcal{L}(D_i, &\theta_{i,b_{1}}^{(t_2)}, \theta_{i,b_{2}}^{(t_2)}, \ldots , \theta_{i,b_{K-1}}^{(0)}, \theta_{i,b_{K}}^{(0)}, \theta_{i,h}^{(0)})  \nonumber \\ 
            &\vdots \nonumber \\
            \theta_i^{(t_K+1)} = \theta_i^{(t_K)} - \eta \nabla_{\theta_i} \mathcal{L}(D_i, &\theta_{i,b_{1}}^{(t_K)}, \theta_{i,b_{2}}^{(t_K)}, \ldots , \theta_{i,b_{K-1}}^{(t_L)}, \theta_{i,b_{K}}^{(t_K)}, \theta_{i,h}^{(0)}).
\end{align}
\label{vanilla_equation}
\end{small}

In Eq.~{(5)}, Once the first base layer $\theta_{i,b_1}$ is sufficiently trained, the freeze on the next base layer $\theta_{i,b_2}$ is released for training. This sequential unfreezing and training of base layers up to $\theta_{i,b_K}$ defines Vanilla scheduling. The \{$t_1, t_2, \ldots, t_K \in T$\} represent the global rounds where the freeze of each layer is released. Throughout the training process, the head layer $\theta_{i,h}$ is initialized and kept frozen, only being utilized in the final fine-tuning stage.


\subsection{Method 2: Anti Scheduling}
Anti scheduling is a training approach that initiates with the deepest layers (closest to the output) in the base layer, freezing the remaining layers, and then progressively unfreezing them for sufficient training. This method allows the model to preferentially grasp high-level and various features. The training steps of Anti scheduling, starting with the deepest layers, can be given as


\begin{small}
\begin{align}
            \theta_i^{(t_1+1)} = \theta_i^{(t_1)} - \eta \nabla_{\theta_i} \mathcal{L}(D_i, &\theta_{i,b_{1}}^{(0)}, \theta_{i,b_{2}}^{(0)}, \ldots , \theta_{i,b_{K-1}}^{(0)}, \theta_{i,b_{K}}^{(t_1)}, \theta_{i,h}^{(0)})  
            \nonumber \\
            \theta_i^{(t_2+1)} = \theta_i^{(t_2)} - \eta \nabla_{\theta_i} \mathcal{L}(D_i, &\theta_{i,b_{1}}^{(0)}, \theta_{i,b_{2}}^{(0)}, \ldots , \theta_{i,b_{K-1}}^{(t_2)}, \theta_{i,b_{K}}^{(t_2)}, \theta_{i,h}^{(0)})  \nonumber \\ 
            &\vdots \nonumber \\
            \theta_i^{(t_K+1)} = \theta_i^{(t_K)} - \eta \nabla_{\theta_i} \mathcal{L}(D_i, &\theta_{i,b_{1}}^{(t_K)}, \theta_{i,b_{2}}^{(t_K)}, \ldots , \theta_{i,b_{K-1}}^{(t_K)}, \theta_{i,b_{K}}^{(t_K)}, \theta_{i,h}^{(0)}).
\end{align}
\label{anti_equation}
\end{small}

In Eq~{(6)}, once the deepest base layer $\theta_{i,b_K}$ is sufficiently trained, the freeze on the preceding layer $\theta_{i,b_{K-1}}$ is released for training. Anti scheduling is thus defined as the sequential unfreezing and training of base layers starting from $\theta_{i,b_K}$ and proceeding to the first base layer $\theta_{i,b_1}$. The training progresses in the reverse order of Anti scheduling, and during this process, the head layer $\theta_{i,h}$ remains frozen and is only utilized in the final fine-tuning phase.


\begin{algorithm}[htb!]
\caption{Layer Decoupling Algorithm with Vanilla and Anti Scheduling}
\label{alg:modified_combined_scheduling_algorithm}
    \textbf{Input}: Total global rounds $T$, Total clients $N$, join ratio $r$, Learning rate $\eta$, Total base layers $K$, Fine-tuning rounds $F$, Scheduling Mode $Mode$, Layer Unfreeze Rounds $t_1, t_2, \ldots, t_K$ \\
    \textbf{Initialize}: Global base parameters $\theta_{G,b_{1}}^{(0)}, \theta_{G,b_{2}}^{(0)}, \ldots , \theta_{G,b_{K}}^{(0)}$ and Global Head parameter $\theta_{G,h}^{(0)}$ \\
    \begin{algorithmic}[1] 
    \FOR{$t = 1$ \TO $T$}
        \STATE Randomly select $M = \lfloor r \times N \rfloor$ clients
        \FOR{Each selected client $i = 1$ \TO $M$}
            \IF{$Mode = \text{'Vanilla'}$}
                \IF{$t \geq t_{k}$}
                    \STATE Unfreeze $\theta_{i,b_{k}}$
                \ENDIF
                \STATE $\theta_{i}^{(t+1)} = \theta_i^{(t)} - \eta \nabla_{\theta_{i}} \mathcal{L}(D_i,\theta_{i,b_{1}}^{(t)}, \ldots, \theta_{i,b_{K}}^{(t)}, \theta_{i,h}^{(0)})$
            \ELSIF{$Mode = \text{'Anti'}$}
                \IF{$t \geq t_{k}$}
                    \STATE Unfreeze $\theta_{i,b_{K-k+1}}$
                \ENDIF
                \STATE $\theta_{i}^{(t+1)} = \theta_i^{(t)} - \eta \nabla_{\theta_{i}} \mathcal{L}(D_i,\theta_{i,b_{1}}^{(t)}, \ldots, \theta_{i,b_{K}}^{(t)}, \theta_{i,h}^{(0)})$
            \ENDIF
            \STATE Keep $\theta_{i,h}$ frozen
            \STATE Send updated base parameters $\theta_{i,b_{1}}^{(t+1)}, \ldots, \theta_{i,b_{K}}^{(t+1)}$ 
        \ENDFOR
        \STATE Aggregate global model $\theta_G^{(t+1)} = \sum_{i=1}^{N} \frac{|D_i|}{|D|} \theta_{i,b}^{(t+1)}$
    \ENDFOR

    \FOR{$f = 1$ \TO $F$} 
        \STATE Unfreeze all layers for each client
        \FOR{each client $i = 1$ \TO $N$}
            \STATE Update all parameters $\theta_{i}$ for fine tuning
            \STATE $\theta_{i}^{(T+f)} = \theta_{i}^{(T+f-1)} 
            \qquad \qquad \qquad - \eta \nabla_{\theta_{i}} \mathcal{L}(D_i,\theta_{i,b_{1}}^{(T+f-1)}, \ldots, \theta_{i,b_{K}}^{(T+f-1)}, \theta_{i,h}^{(T+f-1)})$
        \ENDFOR
    \ENDFOR
\end{algorithmic}
\end{algorithm}

Algorithm~\ref{alg:modified_combined_scheduling_algorithm} outlines the training procedure for both Vanilla and Anti scheduling. Line~2 represents client sampling. Lines~4--8 represent the Vanilla scheduling mode. When the Vanilla mode is selected, if the current global round $t$ is greater than the layers unfreeze round $t_k$, as specified in line~5, the algorithm unfreezes the $k$-th base parameter $\theta_{i,b_k}$. Then, in line~8, the unfrozen base parameters are used to perform a local update. 

Lines~9--13 describe the Anti scheduling mode. In Anti mode, as line~10 indicates, if the current global round $t$ exceeds the layer unfreeze round $t_k$, the algorithm unfreezes the ($K-k+1$)-th base parameter $\theta_{i,b_{K-k+1}}$, which is closer to the head layer. Line~13 then uses the unfrozen base parameters for the local update. Lines~15--16 maintain the freezing of the head while sending the updated base parameters to the server. Line 18 performs global aggregation after local updates, and lines~20--24 is the fine-tuning process in which both base and head parameters are utilized in each client after the global rounds are complete



\section{Experiments}
\begin{figure}[htb!]
\centering
\includegraphics[width=0.5\textwidth]{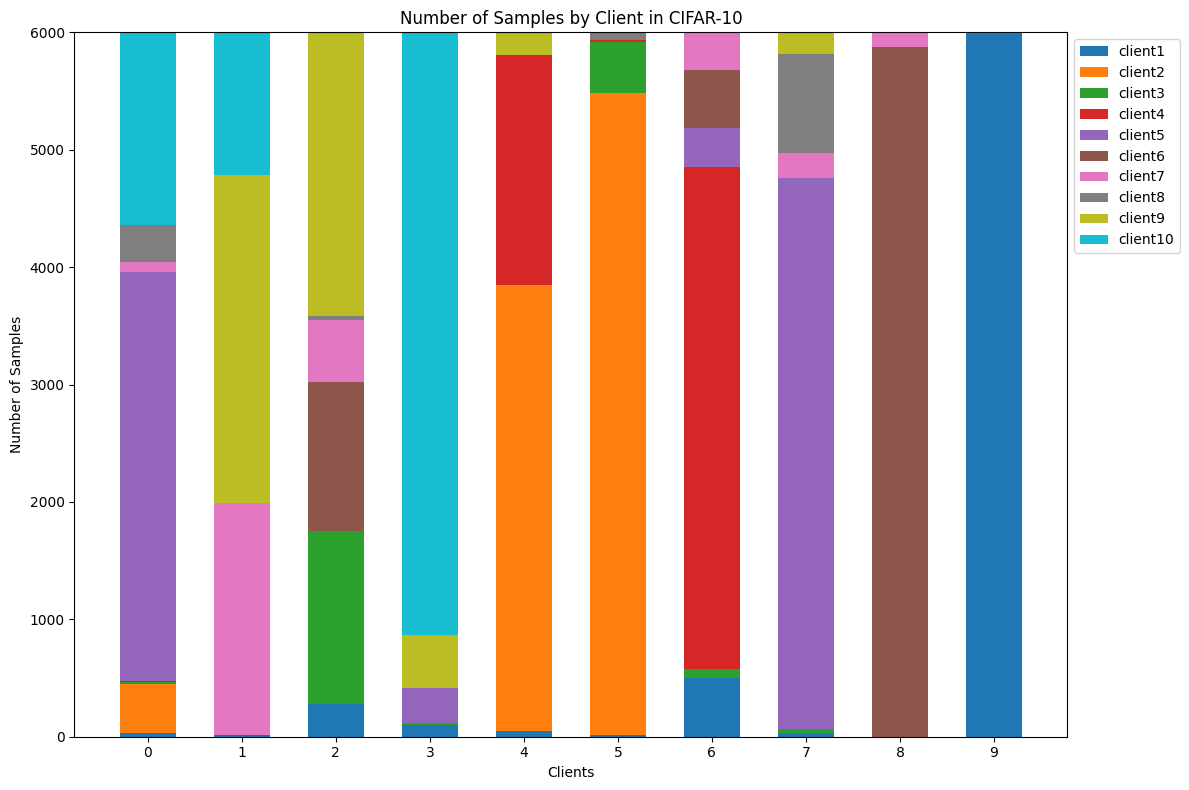}
\caption{Results of sampling the CIFAR-10 dataset among 10 clients from a Dirichlet distribution. The Dirichlet parameter $\alpha$ is 0.1, generating highly heterogeneous data. However, the nature of the CIFAR-10 dataset does not offer much class heterogeneity.}
\label{data_infomation}
\end{figure}

In our experimental setup, the total number of clients $N$ is set to 100, with each round involving a client participation ratio $r$ of 0.1, indicating that 10 clients participate in each training round. The batch size is configured to 10, and the learning rate is set at 0.005. The datasets used in our experiments include MNIST, CIFAR-10, CIFAR-100, and Tiny-ImageNet. To induce heterogeneity among the data distributed to each client, we sampled from a Dirichlet distribution with a Dirichlet parameter of 0.1, establishing a highly heterogeneous environment. This approach to data distribution is visualized in Figure~\ref{data_infomation}, which illustrates the results of sampling the CIFAR-10 dataset among 10 clients. Despite the high heterogeneity induced by the Dirichlet parameter $\alpha$ of 0.1, MNIST and CIFAR-10 dataset inherently lacks significant class heterogeneity. Therefore, our experiments focus on CIFAR-100 and Tiny-ImageNet datasets because our approach performs well in environments with significant data and class heterogeneity. The CIFAR-100 dataset contains 100 different classes, and the Tiny-ImageNet dataset contains 200 classes.


In our experiments, we compare our approach against six baselines, as detailed in the following references: ~\cite{arivazhagan2019federated,chen2021bridging,collins2021exploiting,liang2020think,mcmahan2017communication,oh2021fedbabu}. The model used is a shallow CNN model comprising two convolutional layers and two fully connected layers. Here, the last fully connected layer is set as the \textit{head}, and the remaining three layers are designated as the base. The total number of base layers $K$ is set to 3, with $t_1$ at the 0 (initial) round, $t_2$ at the 100-th global round, and $t_3$ at the 200-th global round. Additionally, in FedBABU~\cite{oh2021fedbabu}, the learning rate of the \textit{head} is set to zero during the training rounds, implying that gradients are calculated but not applied. In contrast, our scheduling algorithm achieves complete decoupling by freezing the gradients, thus preventing both computation and application. We will delve into the comparison of computational and communication costs in Section~\ref{computation cost}.


\begin{table}[htb!]
\caption{Comparison of accuracy}
\resizebox{\columnwidth}{!}{%
\begin{tabular}{c|cc|cc|cc|cc}
\hline
\multirow{2}{*}{Algorithm} & \multicolumn{2}{c|}{MNIST} & \multicolumn{2}{c|}{CIFAR-10} & \multicolumn{2}{c|}{CIFAR-100} & \multicolumn{2}{c}{Tiny-ImageNet} \\ \cline{2-9} 
                        & Acc. (\%) & $T$ & Acc. (\%) & $T$ & Acc. (\%) & $T$ & Acc. (\%) & $T$ \\ \hline
\multirow{3}{*}{FedAvg~\cite{mcmahan2017communication}}           
& 91.48 & 100 & 36.87 & 100 & 29.04 & 500 & 16.97 & 300           \\
& 96.32 & 200 & 42.00 & 200 & 30.95 & 1000 & 14.06 & 500           \\
& 97.48 & 300 & 49.36 & 300 & 30.34 & 1500 & 14.27 & 1500           \\
\hline
\multirow{3}{*}{FedPer~\cite{arivazhagan2019federated}}
& 98.30 & 100 & 86.16 & 100 & 40.22 & 500 & 29.04 & 300           \\
& 98.63 & 200 & 87.34 & 200 & 41.03 & 1000 & 29.24 & 500           \\
& 98.69 & 300 & 86.66 & 300 & 40.05 & 1500 & 29.11 & 1000           \\
\hline
\multirow{3}{*}{LG-FedAvg~\cite{liang2020think}}
& 97.95 & 100 & 85.75 & 100 & 41.06 & 500 & 26.46 & 300           \\
& 98.17 & 200 & 86.10 & 200 & 41.01 & 1000 & 26.35 & 500           \\
& 98.13 & 300 & 85.91 & 300 & 40.80 & 1500 & 26.36 & 1000           \\
\hline
\multirow{3}{*}{FedRep~\cite{collins2021exploiting}}
& 98.56 & 100 & 86.58 & 100 & 40.44 & 500 & 28.85 & 300           \\
& 98.69 & 200 & 88.12 & 200 & 41.24 & 1000 & 28.54 & 500           \\
& 98.75 & 300 & 88.11 & 300 & 40.66 & 1000 & 28.50 & 1000           \\
\hline
\multirow{3}{*}{FedROD~\cite{chen2021bridging}}
& 98.61 & 100 & 86.24 & 100 & 49.50 & 500 & 37.23 & 300            \\
& 98.95 & 200 & 88.01 & 200 & 48.70 & 1000 & 35.34 & 500           \\
& \textbf{99.06} & 300 & 88.59 & 300 & 50.56 & 1500 & 35.35 & 1000           \\
\hline
\multirow{3}{*}{FedBABU~\cite{oh2021fedbabu}}           
& 98.19 & 100 & 85.77 & 100 & 49.28 & 500 & 37.84 & 300           \\
& 98.77 & 200 & 86.16 & 200 & 52.75 & 1000 & 33.78 & 500           \\
& 98.84 & 300 & 87.07 & 300 & 51.82 & 1500 & 32.57 & 1000           \\
\hline
\multirow{3}{*}{\textbf{Vanilla Scheduling}} 
& 98.21 & 100 & 85.82 & 100 & 56.86 & 500 & 41.86 & 300   \\ 
& 98.84 & 200 & 87.81 & 200 & 59.52 & 1000 & 39.51 & 500   \\ 
& 98.99 & 300 & 88.58 & 300 & 58.49 & 1500 & 36.44 & 1000   \\ 
\hline
\multirow{3}{*}{\textbf{Anti Scheduling}} 
& 98.34 & 100 & 85.73 & 100 & 56.17 & 500 & \textbf{41.94} & 300   \\ 
& 98.81 & 200 & 87.67 & 200 & \textbf{60.06} & 1000 & 39.67 & 500   \\ 
& 98.96 & 300 & \textbf{88.61} & 300 & 58.58 & 1500 & 36.63 & 1000   \\ 
\hline
\end{tabular}%
}
\label{comparison_of_accuracy}
\end{table}

Table~\ref{comparison_of_accuracy} presents a comparison of the accuracies achieved by various federated learning algorithms across multiple datasets, including MNIST, CIFAR-10, CIFAR-100, and Tiny-ImageNet. It encompasses representation learning algorithms such as FedAvg, FedPer, LG-FedAvg, FedRep, FedROD, and FedBABU, as well as our proposed Vanilla and Anti scheduling algorithms. A key aspect of both FedBABU and our scheduling algorithms is their approach during the training phases leading up to the final global round, denoted as $T$. Characteristically, these algorithms exclusively use the base layers and do not employ the head layer during the training rounds, resulting in relatively lower accuracies prior to the final round. To ensure a fair comparison, the accuracy figures for FedBABU and our scheduling algorithms at $T$ represent the performance after fine-tuning. Additionally, it is noteworthy that our scheduling algorithms have achieved high accuracy in environments with significant data and class heterogeneity, demonstrating their effectiveness in processing complex and diverse datasets. 

\begin{figure}[htb!]
\centering
\includegraphics[width=0.5\textwidth]{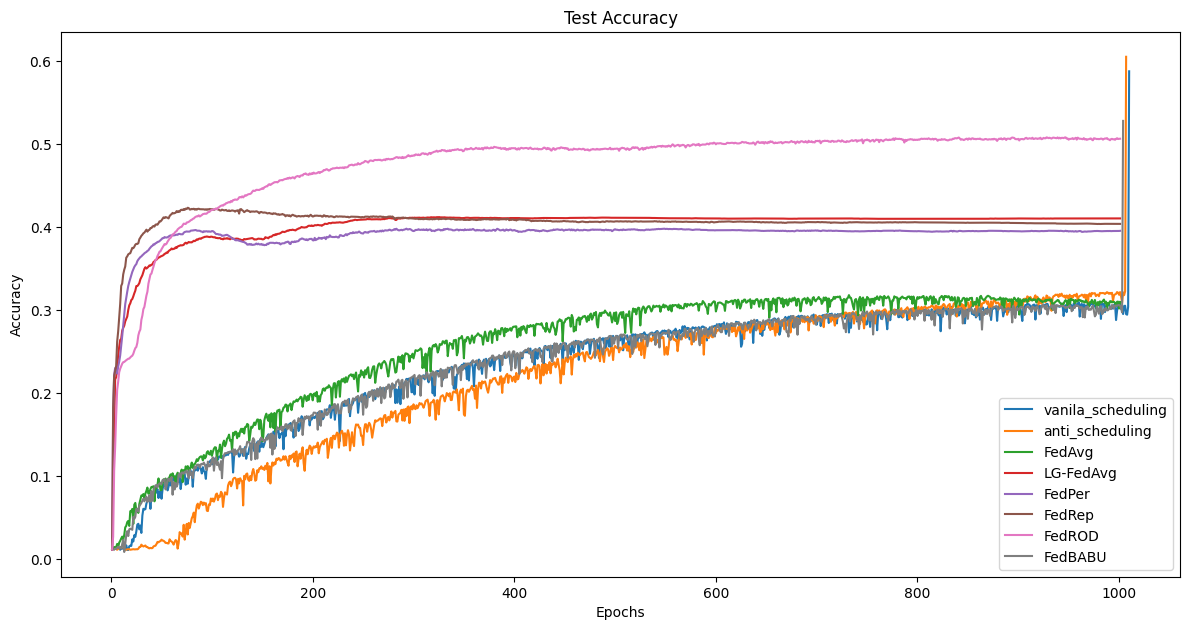}
\caption{This figure shows the average accuracy of clients on the CIFAR-100 dataset. As can be seen, the earlier round accuracy of our scheduling algorithm is lower than FedAvg and FedBABU. This is because, in the earlier rounds, not all base and head participate and the training is conducted using only a portion of the unfrozen base layers.}
\label{Cifar-100}
\end{figure}

Figure~\ref{Cifar-100} shows the average accuracy of clients on the CIFAR-100 dataset, visualizing the performance of our scheduling algorithms compared to baselines. As can be seen, the earlier round accuracy of our scheduling algorithm is lower than FedAvg and FedBABU, which is attributed to the fact that, in the earlier rounds, not only all base but also head participate, and the training is conducted using only a portion of the unfrozen base layers.

\begin{figure}[htb!]
    \centering
    \includegraphics[width=0.5\textwidth]{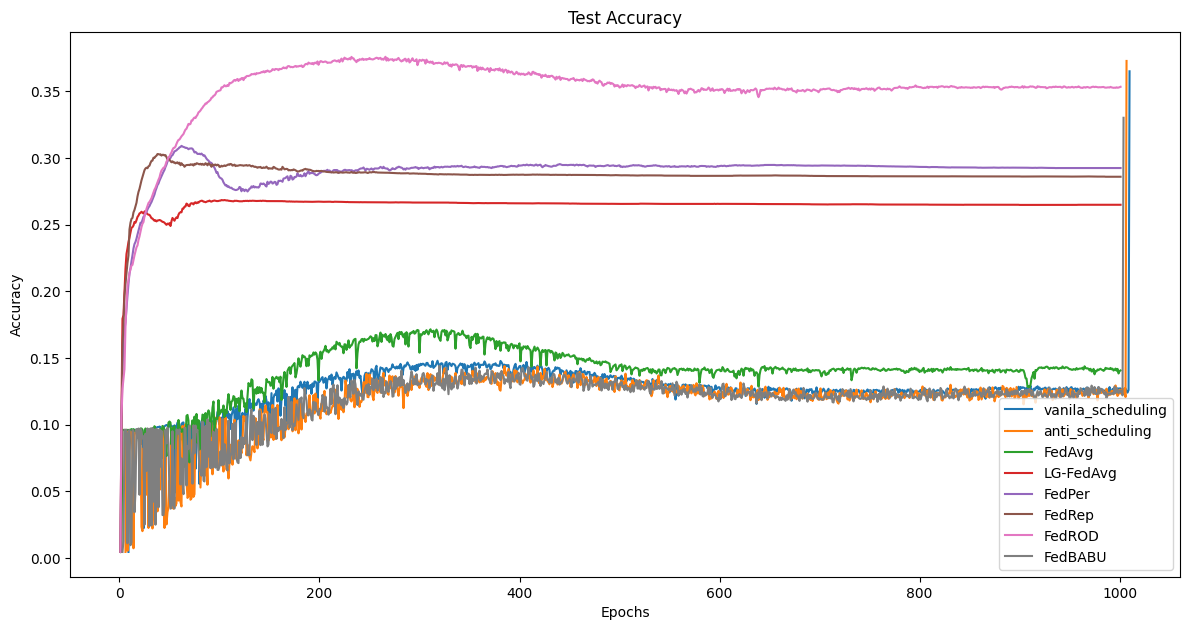}
    \caption{This figure represents the average accuracy of clients on the Tiny-ImageNet dataset, where the initial round accuracy of our scheduling algorithm is lower compared to FedAvg and FedBABU, consistent with its characteristics.}
    \label{Tiny-imagenet}
\end{figure}

Similarly, Figure~\ref{Tiny-imagenet} represents the average accuracy of clients on the Tiny-ImageNet dataset. Our scheduling algorithm's earlier round accuracy is lower than FedAvg and FedBABU, consistent with its characteristics. This visualization helps underscore how our scheduling algorithms gradually catch up and potentially surpass other algorithms as the training progresses.

\section{Ablation Study}

In our ablation study, we demonstrate the results of varying different parameters. The contents to be covered in the ablation study include:
\begin{itemize}
    \item Comparison of client-specific accuracy
    \item Estimation of computational cost for each algorithm
    \item Effect of layer unfreezing timing on the accuracy
    \item Application of scheduling to the baseline algorithms
\end{itemize}


\subsection{Comparison of Client-specific Accuracy}
To ensure that accuracy improvements during fine-tuning are not biased or limited to specific clients, we conducted a comparative analysis of client-specific accuracy between the latest representation learning algorithms and our algorithm. Since the accuracy differences in the MNIST and CIFAR-10 datasets are not significantly pronounced, the comparison was focused on the CIFAR-100 and Tiny-ImageNet datasets. The results demonstrated that the higher average accuracy of our scheduling algorithm was not due to bias towards specific clients but was consistently achieved across all clients.

\begin{figure}[htb!]
\centerline{\includegraphics[width=0.5\textwidth]{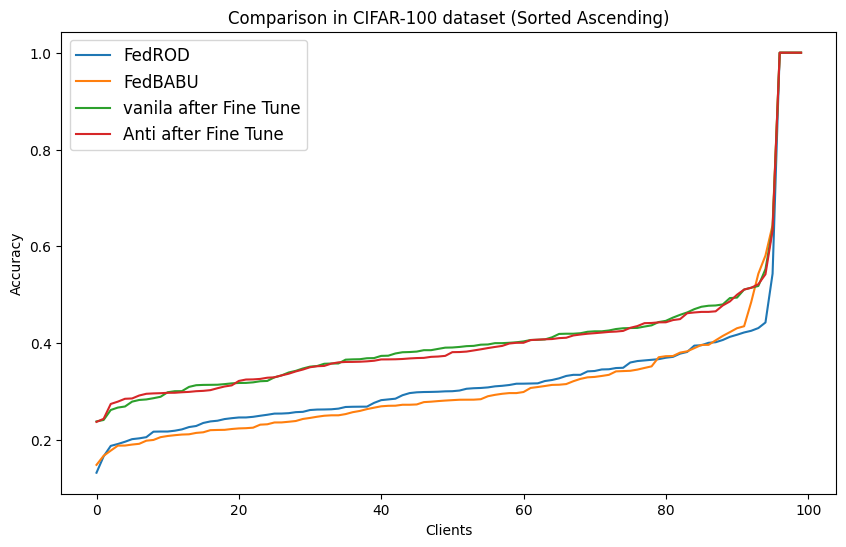}}
\caption{Comparison of client-specific accuracy in the CIFAR-100 dataset. This figure visualizes the accuracy of each client in ascending order.}
\label{CIFAR-100}
\end{figure}

Figure~\ref{CIFAR-100} shows the comparison of client-specific accuracy in the CIFAR-100 dataset. This visualization helps to highlight that our scheduling algorithm achieves consistent accuracy across a spectrum of different clients without favoring any particular group.

\begin{figure}[htb!]
\centerline{\includegraphics[width=0.5\textwidth]{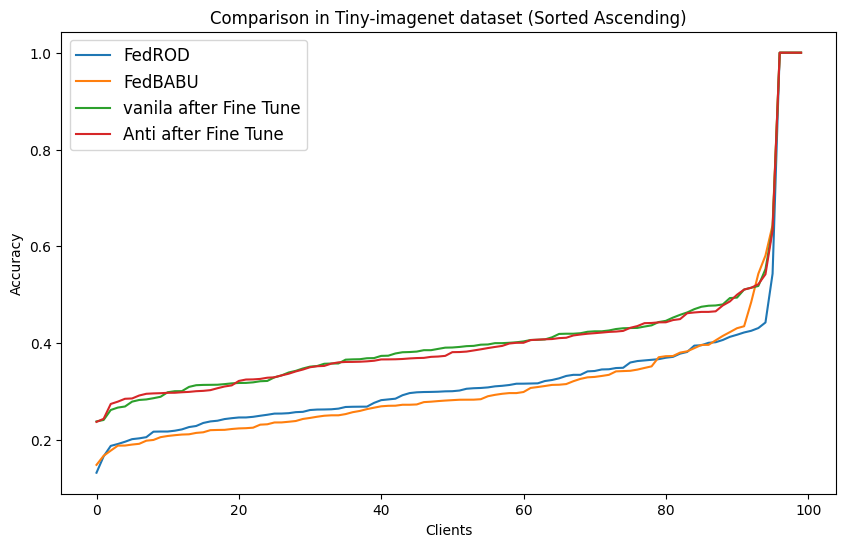}}
\caption{Comparison of client-specific accuracy in the Tiny-ImageNet dataset. This figure visualizes the accuracy of each client in ascending order.}
\label{tiny-imagenet}
\end{figure}

Similarly, Figure~\ref{tiny-imagenet} demonstrates the client-specific accuracy in the Tiny-ImageNet dataset. As with CIFAR-100, the visualization confirms that our algorithm manages to maintain high accuracy uniformly across different clients, showcasing its robustness.

\subsection{Estimation of Computational Cost for Each Algorithm}\label{computation cost}
In the actual experimental environment, using non-IID data makes it challenging to estimate computational costs. Therefore, we estimate the computational costs using the number of FLOPs. The number of parameters in each layer is mentioned following in Table \ref{table:numberofparameters}.


\begin{table}[htb!]
\centering
\caption{Number of Parameter per Layer} 
\begin{tabular}{l|l}
\hline
\multicolumn{1}{c|}{Layer}            & \multicolumn{1}{c}{\# of Parameters} \\ \hline
conv1.weight     & 800                                  \\
conv1.bias       & 32                                   \\
conv2.weight     & 51,200                                \\
conv2.bias       & 64                                   \\
fc1.weight       & 524,288                               \\
fc1.bias         & 512                                  \\
fc2.weight       & 5,120                                 \\
fc2.bias         & 10                                   \\ \hline
Total Parameters & 582,026 \\ \hline
\end{tabular}
\label{table:numberofparameters}
\end{table}


To compare the computational costs of FedAvg, FedBABU, and our scheduling algorithms, we consider the data processed by each client in an IID (independent and identically distributed) environment. Costs are measured in FLOPs, and for simplicity, all algorithms are assumed to use datasets with an equal pixel count per image. In datasets like MNIST, CIFAR-10, and CIFAR-100, each with a total of 50,000 training data and 100 clients, each client processes 500 data points. In the setting of our experiments, a batch size of 10 is used, with each client handling 50 batches per epoch.


For FedAvg, as seen in the Table~\ref{table:numberofparameters}, the total model parameters are 582,026, and the computational cost per round is $582,026 \times 50$. Thus, the total computational cost for all clients and all rounds is 873.039 billion FLOPs. In the same environment, the computational cost for FedBABU, considering that the \textit{head} (fc2) layer parameters are not computed during the training rounds, results in 576,896 learning parameters. Hence, the total computational cost for FedBABU is 865.344 billion FLOPs. Using the same method for our scheduling algorithms, the computed parameters result in 314.912 billion FLOPs for Vanilla Scheduling and 838.880 billion FLOPs for Anti Scheduling, as detailed in Table \ref{table4}.

\begin{figure}[htb!]
\centerline{\includegraphics[width=0.4\textwidth]{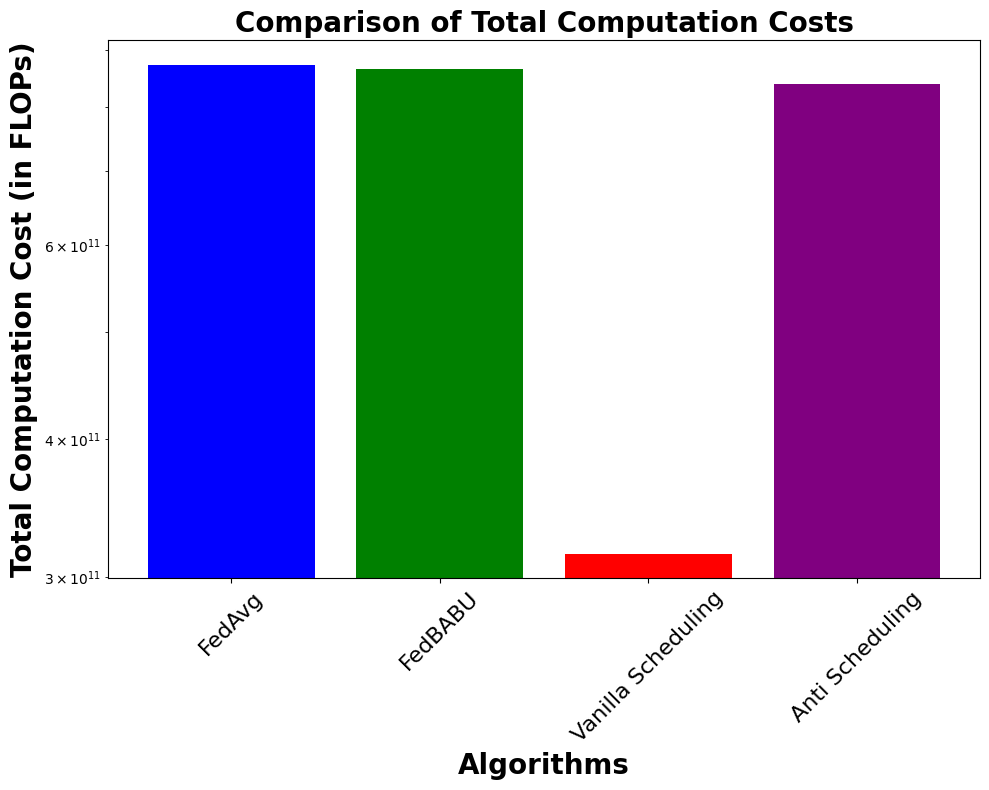}}
\caption{Comparison of computational costs between FedAvg, FedBABU, and our scheduling algorithms. Vanilla scheduling significantly reduces computational costs in the initial rounds by updating only the front-end layers.}
\label{Computation_cost}
\end{figure}

Figure~\ref{Computation_cost} visually compares the computational costs between FedAvg, FedBABU, and our scheduling algorithms. It illustrates how Vanilla scheduling significantly reduces computational costs in the earlier rounds by updating only the earlier unfreezed layers. This visualization supports the data presented in Table \ref{table4}, emphasizing the efficiency of Vanilla Scheduling in managing computational resources.


\begin{table}[htb!]
\centering
\caption{Comparison of Computational Cost}
\begin{tabular}{c|cc}
\hline
Algorithm & \# of FLOPs $(\times10^9)$ & $T$ \\ \hline
FedAvg~\cite{mcmahan2017communication}           
& 873,04 & 300     \\ 
FedBABU~\cite{oh2021fedbabu}
& 865,34 & 300      \\ 
Vanilla Scheduling
& 314,91 & 300       \\  
Anti Scheduling
& 838,88 & 300        \\  \hline
\end{tabular}%

\label{table4}
\end{table}

\subsection{Accuracy Differences Due to Changes in Layer Unfreezing Timing}
In our two scheduling algorithms, the number of base layers, $t_k$, is set to 3, leading to parameter unfreezing occurring across three phases. For instance, in Vanilla scheduling, training proceeds with the unfreezing of the conv1 layer from round $T=0$ to 100. Subsequently, from rounds 100 to 200, both conv1 and conv2 layers are unfrozen for training. Finally, from round 200 to the final global round, all base layers, including conv1, conv2, and fc1, are unfrozen and utilized for training. While the unfreezing points in this paper are arbitrarily set for discussing the effectiveness of layer decoupling, an appropriate scheduling method is required accordingly. Therefore, we compare the differences in outcomes when unfreezing occurs at rounds 50 and 100 instead of 100 and 200.


In the CIFAR-100 dataset, with the global round $T=1000$ and layer unfreezing round set at $t_1=0, t_2=100, t_3=200$, Vanilla scheduling achieves an accuracy of 59.52\%, while Anti scheduling achieves 60.06\%. However, when the unfreezing round is changed to $t_1=0, t_2=50, t_3=100$, Vanilla scheduling achieves 58.68\%, and Anti scheduling achieves 59.02\% accuracy, slightly lower accuracy. For the Tiny-ImageNet dataset at global round $T=300$, with $t_1=0, t_2=100, t_3=200$, Vanilla and Anti scheduling achieve accuracies of 41.86\% and 41.94\%, respectively. Changing the unfreezing points to $t_1=0, t_2=50, t_3=100$ results in accuracies of 41.23\% for Vanilla and 41.63\% for Anti scheduling, which are also slightly lower. The insight gained here is that while the timing of layer unfreezing does not significantly affect accuracy, as seen in Section~\ref{computation cost}, it has a substantial impact on computational cost. Therefore, setting larger values for $t_k$ is advantageous where possible.


\subsection{Application of Scheduling to Baseline Algorithms}
When our scheduling methods are applied to the baseline algorithms, there was no observed improvement in accuracy. This lack of improvement is attributed to the fact that, except for~\cite{oh2021fedbabu}, other algorithms update the head during training rounds, which nullifies the effect of freezing base layers. Furthermore, Vanilla scheduling achieved a performance similar to the non-scheduled approach, while Anti scheduling, in particular, resulted in significantly lower accuracy. Therefore, natively applying a dense division of the base layer only for scheduling purposes might inadvertently lead to a negative impact on accuracy. Additionally, if the head layer is utilized for local updates during training, the intended effect of freezing the base layers is effectively nullified. 




\section{Conclusion}
This research introduces a layer decoupling technique in representation learning, a domain of personalized federated learning, to address data and class heterogeneity among clients. In traditional representation learning, the model is divided into 'base' and 'head' components, each customized to process the common and unique features of client data, respectively. Our primary contribution focuses on the dense division of the base component in representation learning, and the subsequent application of two novel scheduling methodologies, namely Vanilla and Anti scheduling, to these finely segmented base layers in federated learning architecture. The Vanilla scheduling approach excels in significantly reducing computational costs without degrading model performance. This aspect is particularly crucial in federated learning environments where resources and computational power are often limited. Conversely, Anti scheduling demonstrates superior capability in handling scenarios with pronounced data and class heterogeneity. Our experimental analysis, conducted on datasets with high data and class heterogeneity, such as CIFAR-100 and Tiny-ImageNet, shows that these scheduling methods not only contribute to a more efficient training process but also ensure fair improvements in accuracy across all clients. This addresses the challenges of personalized federated learning by ensuring that no single client's data disproportionately influences the global model.





\bibliographystyle{unsrt}  








\end{document}